# Automated 3D Kinematic Monitoring for Circadian Activity and Anomaly Detection in Juvenile Fish

Chih-Wei Huang[1*], Chang-Wen Huang[2,3], Chung-Ping Chiang[3], Tsung-Wei Pan[1]

[1] AI Research Center, National Taiwan Ocean Univ., Keelung City 20224, Taiwan.
[2] Dept. of Aquaculture, National Taiwan Ocean Univ., Keelung City 20224, Taiwan.
[3] Center of Excellence for the Oceans, National Taiwan Ocean University, Keelung City 20224, Taiwan.
* Corresponding author: Jung-Hua Wang (jhwang@email.ntou.edu.tw)

*Abstract*—**Precision aquaculture faces a "phenotyping bottleneck" in tracking high-resolution behavioral traits, as conventional methods cannot quantify instantaneous three-dimensional (3D) physical exertion. To address this, we present a high-throughput 3D behavioral phenotyping framework integrating deep learning object detection with binocular stereo vision for real-time monitoring of juvenile tilapia in high-density environments. The system automates non-contact body length estimation and reconstructs 3D swimming trajectories from absolute spatial coordinates. By eliminating 2D perspective distortions, this approach precisely quantifies 3D velocity and acceleration, marking the first estimation of true physical swimming speeds in free-roaming juveniles. Results show the framework successfully establishes circadian locomotor baselines, serving as an early warning system for physiological stress and providing an objective metric for fish vitality.**

***Keywords—Deep learning, Juvenile fish activity, Phenotypic trait, Smart breeding, Stereo vision, Tilapia***

## I. Introduction

Aquaculture is a cornerstone of global food security, with selective breeding serving as a pivotal strategy to enhance production resilience [1]. While modern genomic tools enable high-yielding, disease-resistant strains, evaluating these genetic gains requires precise quantification of complex morphological and behavioral traits—moving beyond traditional, stressful manual measurements of weight and length [2].

However, the phenotypic expression of species like Tilapia is highly sensitive to environmental variables like water quality, temperature, and stocking density [3]. This sensitivity creates a **"phenotyping bottleneck,"** where rapid genomic data expansion far outpaces the availability of high-resolution phenotypic tracking [4]. Because static, single-point measurements cannot capture dynamic responses to environmental stressors, longitudinal data with fine temporal resolution is essential to deciphering how different genotypes manifest their growth potential under fluctuating conditions.

Monitoring juvenile fish (aged 1–4 months) optimizes data stability by minimizing maturity-related biological noise, yet large-scale data acquisition remains technically challenging. Conventional manual capture induces physiological stress and immunosuppression [5]. Meanwhile, intensive indoor breeding environments present high stocking densities, severe inter-object occlusions, and rapid, non-rigid pose variations that hinder precise observation.

Current non-contact ichthyometrics offer automated biometric estimations [6-8], such as AIoT frameworks for static body length measurement [6]. However, high-frequency kinematic tracking remains an underdeveloped frontier. Existing smart aquaculture frameworks generally focus on macro-level AIoT system integration or 2D behavioral classification, which lack the spatial depth needed to compute absolute 3D velocities. Conversely, 3D trajectory tracking systems designed for open-ocean settings (e.g., Atlantic salmon research [9]) are optimized for large adult fish and fail to quantify the instantaneous 3D physical exertion of free-roaming juveniles in high-stocking indoor tanks.

To bridge the gap between high-speed visual detection and precise physical measurement, this study integrates a deep learning object detector (e.g., YOLO) with binocular stereo vision. Building on the body length model from [6], our framework maps 2D bounding box centroids into stereoscopic 3D space to reconstruct absolute spatial coordinates, completely eliminating 2D perspective distortions.

This synergistic approach automates the acquisition of high-fidelity locomotor data, specifically 3D swimming velocity and acceleration. To our knowledge, this represents the first attempt to estimate the true physical swimming speeds of free-roaming juvenile fish by building upon established ichthyometric models. By establishing a non-interruptive circadian baseline of locomotor rates, this framework automates health monitoring and acts as an early warning system for physiological anomalies—providing a more direct, objective metric for fish vitality than existing growth-oriented or 2D-based models.

## II. Related Work

### A. Deep Learning-based Object Detection

While two-stage deep learning-based detectors such as Region-based Convolutional Neural Networks (R-CNN) and Transformer-based models [10] offer high accuracy, their latency is prohibitive for edge devices. Thus, the one-stage object detector achieves an optimal equilibrium between speed and precision, delivering the rapid processing required for high-density, dynamic underwater monitoring [11]. This critical need to balance multi-stream feature fusion with strict edge-computing hardware constraints is a prominent challenge across diverse domains, mirroring recent edge-AI methodologies that deploy end-to-end deep fusion and localized landmark visualization on restricted hardware (e.g., [12]. For the specific task of activity level estimation, highly stable object localization is a fundamental prerequisite."

For the specific task of activity level estimation, highly stable object localization is a fundamental prerequisite. While this study adopts a lightweight one-stage object detector (i.e., YOLO11s) to generate robust bounding boxes, the proposed framework is fundamentally detector-agnostic and can flexibly integrate any modern architecture. Instead of relying on complex pixel-level semantic segmentation, the framework directly utilizes the geometric centroids of these high-confidence bounding boxes as stable spatial anchors. This strategy effectively localizes the fish in the 2D image plane, providing a reliable and computationally efficient foundation for subsequent 3D stereoscopic tracking and instantaneous velocity calculations.

### B. Object Detection and Stereo Vision

To quantify the swimming kinematics and activity levels of fish, temporal tracking within a 3D spatial domain is a fundamental prerequisite. Recent advancements in deep learning, particularly the YOLO series coupled with Multi-Object Tracking (MOT) algorithms, have demonstrated robust performance in localizing and tracking multiple fast-moving targets in dynamic aquatic environments. However, 2D planar trajectories inherently lack the spatial depth necessary for calculating absolute biophysical metrics, such as instantaneous velocity. Consequently, binocular stereo vision, specifically utilizing algorithms like Semi-Global Block Matching (SGBM) [13], is widely integrated to compute disparity maps and reconstruct 3D coordinates [14].

Despite its utility, extracting stable 3D trajectories for free-roaming juvenile fish encounters unique bio-optical challenges. The body surfaces of fish fry are weakly textured, and their distal fins are semi-transparent, which frequently induces correspondence ambiguities during SGBM processing. When combined with severe inter-object occlusions in high-density tanks, these visual artifacts generate high-frequency bounding box jitters and erroneous depth jumps. In the context of kinematic analysis, these spatial discontinuities manifest as "phantom movements" in the 3D trajectory even when the fish is resting, thereby necessitating robust downstream temporal filtering to prevent the systematic overestimation of activity levels.

### C. Activity Level Estimation

Computer vision-based monitoring of swimming kinematics is a vital non-invasive method for quantifying metabolic health, satiation, and environmental stress in aquaculture [15] . Although multi-object tracking (MOT) [16] is commonly applied to establish swimming trajectories, raw 3D trajectory data is often compromised by stochastic noise. High-frequency detection jitter and depth estimation errors frequently induce "phantom movements" during periods of fish quiescence, leading to a significant overestimation of actual swimming intensity.

Conventional smoothing techniques, such as moving averages or Kalman filters, often fail to properly capture the distinctive "burst-and-coast" swimming patterns of juvenile fish. These traditional methods face an inherent trade-off: excessive smoothing suppresses noise but also attenuates the high-frequency signal components associated with rapid turns and sprints, thereby obscuring biologically significant behaviors. To overcome these limitations, this study utilizes a dual-stage refinement framework integrating a statistical trimmed-mean filter with a spatial "standstill" state decision mechanism. This standstill gate forces the instantaneous 3D velocity to zero when the displacement of the 2D bounding box falls below a predefined threshold, successfully suppressing stationary noise while preserving high-dynamic motion fidelity. Finally, to ensure biological plausibility, an empirical upper limit is established based on the critical swimming speed ($U_{crit}$) of Nile tilapia under optimal thermal conditions [17]. By adopting this physiological boundary condition, the framework robustly differentiates between genuine biological acceleration and systemic measurement errors.

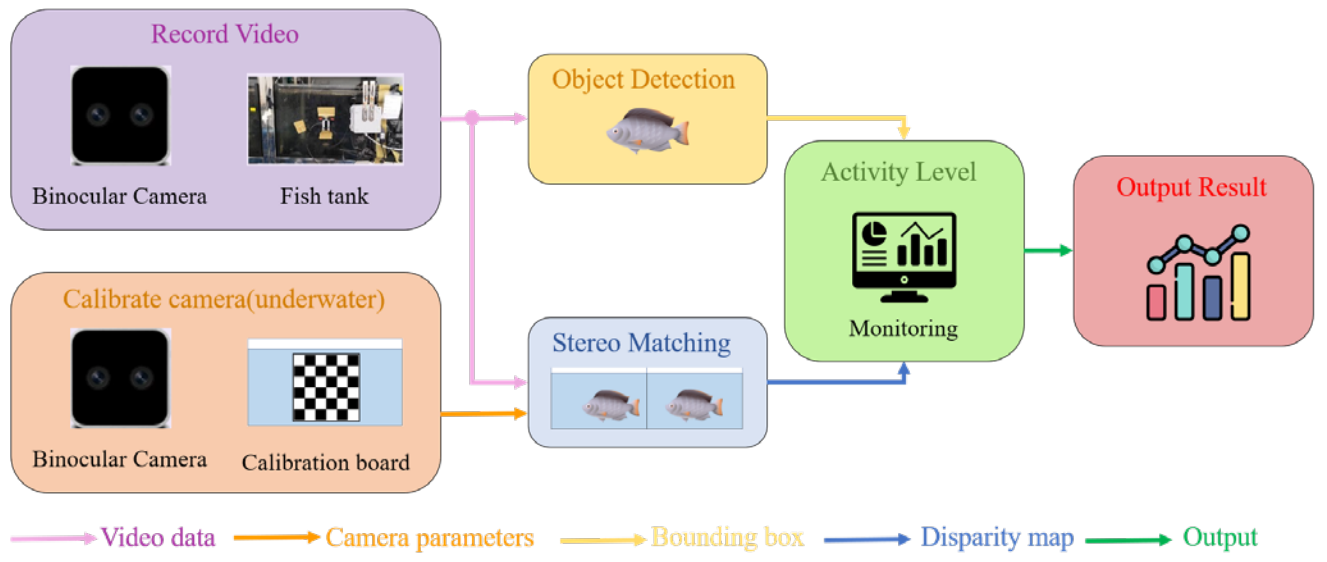

Fig 1. Flowchart of the system for kinematic monitoring fish activity and anomaly detection in juvenile fish

## III. Methodology

### A. Architecture Overview

Fig. 1 illustrates the flowchart of the proposed high-throughput phenotypic measurement system, a unified framework integrating computer vision with stereo geometry to estimate behavioral traits. The workflow initiates with data acquisition using a binocular camera, where underwater calibration is applied to mitigate refractive distortion and obtain precise camera parameters. Subsequently, the video data is processed through two parallel upstream branches sharing the same input source: an object detection module that localizes the fish to generate 2D bounding boxes, and a stereo matching algorithm that computes a dense disparity map. This parallel processing ensures that visual localization and depth information are temporally synchronized for downstream fusion.

Leveraging these shared spatiotemporal features, the system proceeds to the behavioral estimation phase. The detected 2D bounding boxes are mapped onto the 3D disparity space, where positional and depth data are analyzed to evaluate the fish's activity level based on reconstructed swimming trajectories. Ultimately, these behavioral metrics can be aggregated with environmental data to perform genotype-by-environment (G×E) analysis.

### B. Object Detection and Stereo Vision

To acquire reliable 3D swimming trajectories, the system synchronizes 2D target localization with binocular depth estimation. First, the visual data is captured using a stereo depth camera configured for close-range measurements. To overcome the complex light refraction at the water-glass-air interface, an in-situ calibration strategy using Zhang's algorithm [18] is performed directly within the water tank, effectively encapsulating the refractive index into the camera's intrinsic parameters. Following stereo rectification, the disparity map is computed using the Semi-Global Block Matching (SGBM) algorithm.

Concurrently, an off-the-shelf object detector is employed to localize the fish in the image plane. To prevent bounding box jitter from propagating into the downstream 3D reconstruction, a strict confidence threshold of 0.65 is applied, prioritizing spatial fidelity over marginal detection recall. Furthermore, a tracking algorithm (e.g., ByteTrack) is integrated to associate the bounding boxes across consecutive frames, assigning a unique ID to each individual and establishing 2D motion trajectories.

Rather than relying on intensive keypoint annotations, the system explicitly extracts the geometric centroid of each detected bounding box to represent the fish's spatial location. By mapping these 2D spatial centroids onto the 3D disparity space using the calibrated focal length and baseline, the system dynamically derives the 3D world coordinates ($x$, $y$, $z$) of each tracked subject and appends them to a spatial coordinate buffer ($P_{buf}$). This transformation establishes a rigorous geometric basis for subsequent kinematic analysis.

### C. Activity Level Estimation

The fish activity-level analysis starts with the precise localization of fish within 2D images. For each frame, the chosen detector outputs a 2D bounding box for the fish. To convert these regional detection results into point data suitable for trajectory calculation, the system adopts a sequential processing strategy. First, the geometric centroid of each bounding box is extracted; this centroid serves as the representative position of the fish on the image plane. The 2D coordinates are then used as input for the stereo vision to calculate the disparity value and transform it into a 3D world coordinates $P_k = (x_k, y_k, z_k)$, based on which, the velocity of individual fish is calculated.

However, the raw 3D velocities often contain noise due to depth fluctuations and quantization errors. To mitigate the noise effect in this study, a "standstill" state is defined as the tracked fish being in a near-static status if the displacement of the 2D bounding box centroid falls below a calibrated threshold, which corresponds to a planar displacement ($d_{planar}$) of approximately 1.5 mm and the concurrent depth variation ($d_{depth}$) < 2.0mm, the instantaneous 3D velocity is suppressed to zero. Such thresholding ensures that the system only collects genuine swimming maneuvers, effectively filtering out sensory jitters. Only those gated velocities are processed in the next stage. Note that this 1.5 mm physical threshold is specific to the camera optics and baseline object distance of the current setup, and scales directly with observation distance rather than biological fish size.

To eliminate detection-induced jitter, we propose a sliding-window velocity estimation algorithm. Using a non-overlapping 6-frame window ( any incomplete residual frames will be dropped and restart the 6-frame sampling), the local 3D swimming trajectory starting at frame $i$ (where $P_k$ is the spatial coordinate of the $k$-th frame) is defined as:

$$P_i = \{P_k \mid k = i, i+1, \dots, i+5\} \tag{1}$$

where $i$ represents the starting frame index of the current sliding window. Second, instantaneous velocity calculation. Based on the aforementioned coordinate sequence, the system calculates the displacement between adjacent frames within the window. To facilitate comparisons across individuals and studies, all velocities are normalized to Body Lengths per second (BL/s). The instantaneous velocity $V_k$ between the $k$-th and ($k$+1)-th frames is calculated as follows:

$$v_k = \frac{\|P_{k+1} - P_k\|}{\Delta t \cdot L} = \frac{\sqrt{(x_{k+1} - x_k)^2 + (y_{k+1} - y_k)^2 + (z_{k+1} - z_k)^2}}{\Delta t \cdot L} \tag{2}$$

where ||·|| denotes the Euclidean Norm, representing the straight-line distance in 3D space; $\Delta$ is the time interval between frames; and $L$ is a predefined average body length of the monitored juvenile fish. This step yields a set (denoted as $V_{buf}$) containing 5 instantaneous velocity values. Finally, the final displayed speed function uses averaged values because raw instantaneous velocities are susceptible to outliers caused by minor jitters in the object detection bounding boxes, and taking a direct arithmetic mean may lead to distorted results. Therefore, this study adopts the trimmed mean filtering, whereby the five instantaneous velocities are first sorted in ascending order, then the maximum and minimum values are regarded as potential noise and discarded, and the remaining intermediate values are averaged. Thus, the final velocity, $v_{disp}(i)$, is given as

$$v_{disp}(i) = \frac{1}{3}\sum_{j=2}^{4} sort(\{v_k\}_{k=i}^{i+4})_j \tag{3}$$

where $\{v_k\}_{k=i}^{i+4}$ is the set of velocities generated by Eq. (2), and $sort(\cdot)_j$ represents the order statistic, denoting the $j$-th value in the set after sorting. To quantify the fish's circadian rhythm over a prolonged monitoring period, this study employs a segment-wise aggregation strategy. The monitoring timeline is discretized into $M$ non-overlapped sampling intervals (e.g., 10-minute video segments).

For each sampling interval $k$ ($k = 1, 2, \dots, M$), the proposed algorithm calculates the average activity index$\bar{V}_k$, defined as the arithmetic mean of all valid instantaneous velocities within that interval. This process compresses high-frequency kinematic data into a representative activity metric:

$$\bar{V}_k = \frac{1}{N_k}\sum_{t=1}^{N_k} v_{t,k} \tag{4}$$

where $N_k$ is the total number of valid frames in the $k$-th interval, and $v_{t,k}$ denotes the filtered instantaneous velocity at frame $t$. Importantly, to ensure data integrity, any velocity readings exceeding the physiological burst limit (determined as 78 cm/s based on $L_{max}$ cm and a 10 BL/s limit [17]) are excluded from this calculation. Subsequently, a standard Circadian Activity Baseline $\beta$ is constructed as an ordered sequence of these time-indexed activity metrics:

$$\beta = \{(\tau_k, b_k \mid k = 1, 2, \dots, M\}, \tag{5}$$

where $b_k = \bar{V}_k, \tau_k$ represents the time of day, and $b_k$ serves as the normative activity level for that specific time block. This baseline $\beta$ provides a reference profile for detecting behavioral deviations.

To detect health or environmental anomalies, we compare the new observed sequence $C = \{c_1, c_2, \dots, c_M\}$ against the baseline $\beta$ using a score defined with Root Mean Square Error (RMSE, see Eq. (6)). As prescribed in Eq.(7), an alert triggers if the deviation exceeds $2\sigma_{base}$, effectively distinguishing genuine anomalies from normal biological variability.

$$Score = \sqrt{\frac{1}{M}\sum_{k=1}^{M}(c_k - b_k)^2} \tag{6}$$

$$Alarm_trigger = \begin{cases} True\,(Abnormal), & if\ Score > 2\cdot\sigma_{base} \\ False(Normal), & otherwise \end{cases} \tag{7}$$

```
Pseudo Code for Fish Activity Estimation
Input:
  Stereo Images(IL_raw, IR_raw)
  Camera Params (Intrinsics K, Distortion D_coeff)
  Tracking ID set (IDs)
  Time interval (Δt)
  Body Length (L)
  Burst Limit (U_crit)
  Aggregation Interval (M)
Output:
  Filtered Speed( v_disp ), Activity Baseline ( β ),
  Alarm_trigger
1   // Phase 1: Stereo Vision Pre-processing ---
2   Rectify IL_raw and IR_raw to obtain aligned images
    IR , IL
3   Compute Disparity Map D_map using SGBM
4   // Phase 2: Object Detection & 3D Mapping ---
5   For each tracked fish i in IDs do:
6     Extract 2D geometric centroid (cx_k, cy_k) from the
      Bounding Box at frame k
7     Project c_k(cx_k, cy_k) to 3D space using D_map and
      K, D_coeff -> P_k(x, y, z)
8     Append P_k to the P_buf of fish i
9     //Phase 3: Activity Estimation
10    // Ensure the buffer has accumulated at least 6
      consecutive valid frames
11    If Size (P_buf) < 6:
12      Continue (Accumulating Data)
13    Else:
14      V_buf = Empty Array
15      For j = (k-4) to k:  // Compute 5 instantaneous
        velocities from 6 frames
16        // 1. "Standstill" State Gatekeeper
17        d_planar = Distance
          (c_j , c_{j-1})
18        d_depth = |z_j - z_{j-1}|
19        If d_planar < 1.5mm AND d_depth < 2.0mm:
20          v_j= 0  // Set velocity to zero to filter out
            sensory jitters
21          Else:
22          // Compute true 3D instantaneous velocity
            (Unit: BL/s)
23          distance = Distance(P_j - P_{j-1})
24          v_j = distance / (Δt * L)
25        Append v_j to V_buf
26      // 2. Trimmed Mean Filtering
27      Sort(V_buf) in ascending order
28      Remove the maximum and minimum values
          // Discard potential noise
29      v_disp(k) = Mean(remaining 3 values)          //
          Final smoothed velocity
30      // 3. Physiological Plausibility Check
31      If v_disp(k) > U_crit:
32        Discard v_disp(k) // Exceeds biological limit
33  // Phase 4: Circadian Rhythm Aggregation &
        Anomaly Detection ---
34  // Continuous timeline discretized into M intervals
35  For each sampling interval k (1 to M) do:
36      V̄_k = Arithmetic Mean (all valid v_disp in interval
        k)       // Average activity index
37      Append V̄_k to current observation sequence C
38  // Calculate deviation from the Circadian Activity
      Baseline β
39  Score = Calculate RMSE(C, β) using Eq.(6)
40  σ_base = Standard Deviation of β
41  If Score > 2 * σ_base:
42      Alarm_trigger=True(Abnormal)
43  Else:
44      Alarm_trigger= False (Normal)
```

## IV. Experimental Results

Hardware: CPU/i9-9900 with one NVIDIA GPU/RTX-2080 and RAM/32G. Software: Windows 11, python 3.9.21, Cuda 11.8, cuDNN 8.7.0. YOLOv11s under PyTorch 2.0.1. The average fish length in the exemplar tank $L$ is 6.5 cm.

### *A. Activity Level Estimation*

While the system inherently reconstructs continuous 3D trajectories, this study prioritizes kinematic vitality over spatial displacement. Therefore, instead of presenting 3D trajectory plots, we evaluate physiological health directly through instantaneous swimming speeds derived from these trajectories. For fish activity estimation, eight video clips were used. Furthermore, due to the inherent trade-offs in object detection—specifically the strict confidence threshold applied to prioritize spatial fidelity over marginal recall, combined with frequent inter-object occlusions in the high-density tank—the system does not extract valid bounding boxes from every consecutive frame. Consequently, across the video dataset, approximately 70% of the total recorded frames successfully yielded valid detections and were actively utilized for the subsequent kinematic calculations The instantaneous swimming speed between successfully detected frames was first computed and served as the unfiltered single-point baseline, where the speed between any two sequentially detected frames is directly output without temporal filtering. It is important to note that these frames do not need to be strictly consecutive in the video sequence. For instance, if a fish is successfully detected only in frames 1, 3, 4, 5, and 7 due to temporary occlusion or detection failure, the first speed 'point' is calculated between frames 1 and 3, the second between frames 3 and 4, and so forth. To evaluate the suppression of abnormal speed estimates, we compared the single-point baseline against two trajectory post-processing strategies based on a 5-point window, where each 'point' represents the instantaneous speed between two adjacent valid detections, and a total of five such speed points are accumulated for calculation. Note that the term 'point' used in this kinematic context refers solely to a temporal velocity sample, and the total number of evaluated speed points is denoted as $N_{spd}$.

Note that the 5-point filtering strategies process instantaneous velocities in discrete batches. A single averaged velocity point requires a complete set of five instantaneous speeds (derived from at least six valid detected frames). Consequently, the total number of evaluated points ($N_{spd}$) for the 5-point methods is approximately one-fifth of the Single-point baseline, as incomplete residual frames at the end of tracking sequences are excluded.

The two strategies are: (1) 5-point Simple Moving Average (SMA), which averages the five accumulated speed points; (2) 5-point Trimmed mean, defined as the average of the middle three values after excluding the maximum and minimum within this five-point set. Abnormal samples were identified by visually discontinuous 'phantom displacements' or velocities exceeding the aforementioned physiological upper bound ( $U_crit$ = 78 cm/s). Under the unfiltered condition (Raw), roughly 2% of samples (noted as $N_{abn}$) were classified as abnormal or outlier, indicating that simple frame-to-frame differencing tends to overestimate extreme movements. If tested with the five-point SMA, the ratio decreased to 1.19%. Finally, using the same window size but tested with the trimmed mean, the ratio was further reduced to 0.9%. Detailed statistics for the three methods are summarized in Table 1. In short, SMA can remove a large fraction of high-frequency noise and physiologically implausible high-speed values, while the trimmed mean method can further attenuate residual extreme outliers by 50%, yielding the most stable behavior among the three filtering schemes.

As shown in Table 1, the abnormal sample ratio (calculated as $N_{abn}/N_{spd}$ ) is approximately 2.08% under the unfiltered Single-point condition, indicating that simple frame-to-frame differencing is highly susceptible to detection noise Finally, implementing the 5-point Trimmed mean further reduced the abnormal percentage to just 0.9%. This lowest percentage proves that by discarding the maximum and minimum values within the sliding window, the trimmed mean method can further attenuate residual extreme outliers by more than 50% compared to the raw data, yielding the most stable and biologically accurate kinematic estimation among the three schemes.

Table 1. Abnormal speed samples using different trajectory post-processing strategies

| | $N_{spd}(total\ points)$ | $N_{abn}$ | $Percent(\%)$ |
|---|---|---|---|
| Single-point | 417,533 | 8,684 | 2.08 |
| 5-point SMA | 80,443 | 957 | 1.19 |
| 5-point Trimmed mean | 80,443 | 723 | 0.9 |

Applying the trimmed-mean filter and standstill gate effectively suppresses stereoscopic high-frequency noise. As shown in Fig. 2(b), the filtered speed distribution correctly classifies most stationary or gentle cruising states, peaking near 0 BL/s with a mean routine speed of 1.21 BL/s. Furthermore, the full-range distribution (Fig. 2(a)) confirms that the probability density decays rapidly beyond 3 BL/s, maintaining sensitivity to rare burst swimming events (up to 7 BL/s) without systematic over-estimation..

Building upon the accurate kinematic characterization verified in previous sections, the system's capacity for around-the-clock behavior monitoring was evaluated by constructing a 24-hour activity profile to characterize the living habits of the fish (see Fig. 3). The mean swimming intensity aggregated over each sampling interval (visualized as orange data points) and the trend line smoothed using a Moving Average (MA) technique (visualized as the solid red line; e.g., a 3-point

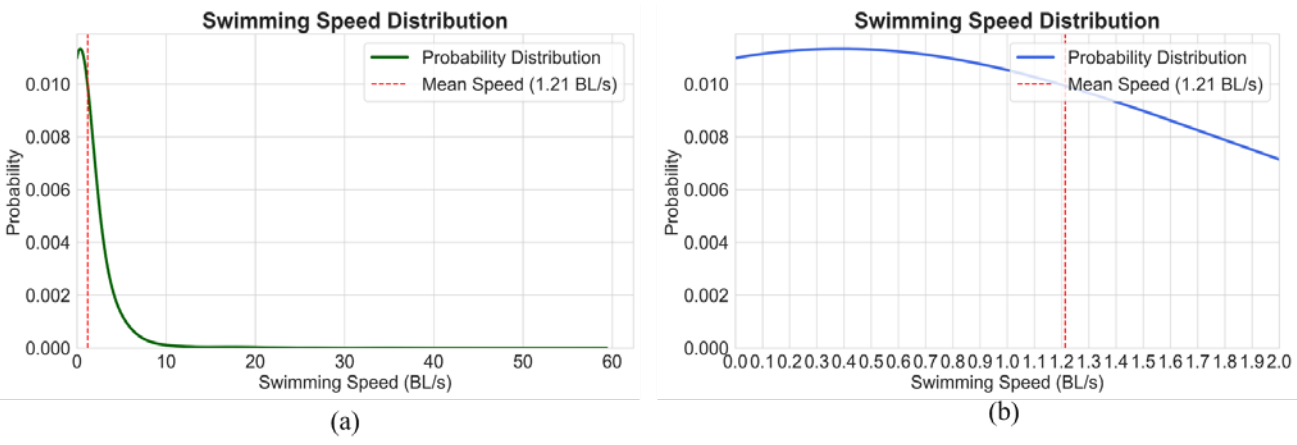

Fig 2. Automated recording data of juvenile tilapia swimming speed of : (a) full-range distribution. (b) zoom-in of the 0-2 BL/s range, with the mean speed (1.21 BL/s).

window was applied here for demonstration) revealed a distinct diurnal rhythm.

Activity levels consistently peaked during the daytime with higher variance, while stabilizing at a low-energy state during the nighttime. This established circadian baseline provides a critical normative reference for behavioral analysis. Unlike short-term observations, this consecutive profile captures the temporal dynamics of the fish's routine behavior. By quantifying deviations from this stable trend (e.g., using the RMSE metric defined in Methodology), the proposed framework offers a quantitative theoretical basis for detecting behavioral anomalies—such as hypoactivity due to disease or hyperactivity due to environmental stress.

To further distinguish between intensity-level anomalies (e.g., lethargy) and pattern-level anomalies (e.g., circadian disruption), we introduce Spearman's Rank Correlation Coefficient $\rho$ as a diagnostic metric. Unlike Pearson's correlation which assumes linearity, Spearman's method evaluates the monotonic relationship based on rank order, making it more robust to non-linear biological variations:

$$\rho = 1 - \frac{6\sum_{i=1}^{M} d_i^2}{M(M^2-1)} \tag{8}$$

where $d_i = rank(x_i) - rank(y_i)$ represents the difference between the ranks of the baseline trend value $x_i$ and the observed activity value $y_i$ at time step $i$, and $M$ is the total number of sampling points.

Unlike Dynamic Time Warping (DTW) [7], which may erroneously align abnormal nocturnal spikes with diurnal peaks, Eq.(8) enforces strict time-point correspondence to detect circadian shifts. Physiologically, $\rho > 0.8$ indicates a preserved rhythm with altered intensity (e.g., fatigue); $\rho \approx 0$ signifies circadian disruption (random activity); and $\rho \approx -1$ implies a severely inverted pattern. Combined with RMSE, this enables a nuanced behavioral diagnosis.

Consequently, the proposed framework demonstrates robust analytical capabilities across multiple temporal scales: from capturing "short-term" high-frequency dynamics (e.g., instantaneous burst swimming), to quantifying "mid-term" activity fluctuations (via segment-wise aggregation), and finally revealing "long-term" behavioral trends (e.g., circadian rhythms). This multi-scale approach provides a comprehensive foundation for precision management in smart aquaculture.

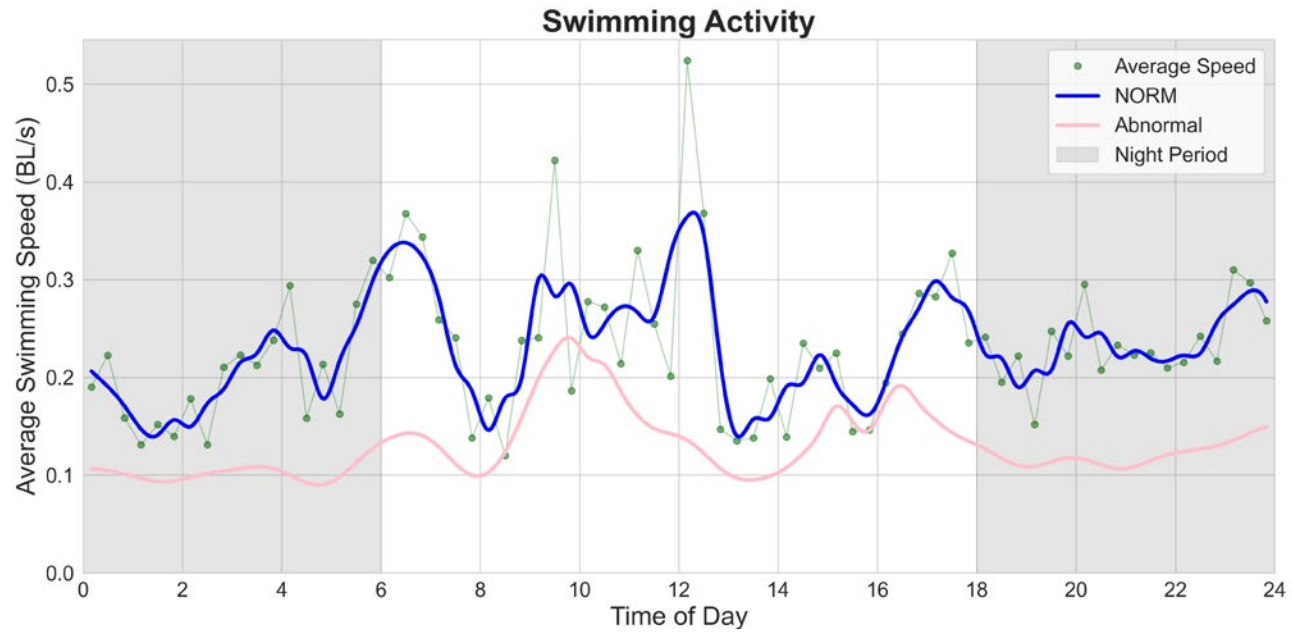

Fig 3. Circadian activity levels derived from swimming kinematic activity, useful for monitoring the health status of fish.

## V. Discussion and Conclusion

The main contribution of this work centers on the development of a non-contact framework dedicated to precise 3D kinematic tracking and activity level estimation in intensive aquaculture. First, we have achieved high-throughput juvenile fish monitoring in high-density environments characterized by complex inter-object occlusions and water turbidity. Second, by extracting 2D geometric centroids from 2D bounding boxes (e.g., via YOLOv11s) and mapping them to a stereoscopic 3D space, the system enables robust trajectory tracking without the constraints of invasive measures. Third, to address the "phantom movements" induced by depth estimation jitter, we proposed a dual-stage kinematic filtering scheme—comprising a standstill gatekeeper and a trimmed-mean filter. This strategy significantly reduced the rate of abnormal velocity samples (from 2.08% to 0.9%) while faithfully preserving the highly dynamic "burst-and-coast" locomotor behaviors. Finally, by aggregating the filtered kinematics and applying a physiological boundary condition ($U_{crit}$), the framework successfully established a 24-hour circadian activity baseline, providing a quantitative basis for automated behavioral anomaly detection.

A pivotal advantage of this work lies in its operational flexibility and spatial fidelity. In contrast to the pixel-level instance segmentation needing computationally intensive annotation or pose-based models requiring labor-intensive keypoint annotations, our method derives precise temporal kinematics directly from high-confidence bounding box centroids coupled with stereo geometry. This design ensures that the derived 3D velocities are physically meaningful (e.g., in BL/s) and effectively neutralizes the systematic overestimation of swimming intensity commonly caused by stereoscopic noise. Moreover, the lightweight nature of the temporal sorting operations required for the trimmed-mean filtering incurs negligible computational overhead, ensuring the system remains resilient for long-term population-level monitoring.

While Semi-Global Block Matching (SGBM) was selected to balance computational load and matching accuracy, disparity computation remains the primary processing bottleneck. This limitation is largely attributed to the current reliance on CPU-based execution, which restricts parallel processing capabilities. To overcome this, future work will explore the integration of state-of-the-art deep learning-based stereo matching architectures and next-generation edge detectors to resolve the current processing bottleneck and achieve real-time performance.

Furthermore, a key operational challenge in high-density tracking is the occurrence of disconnected trajectories when extreme occlusions cause temporary tracking failures. Currently, these broken sequences lead to omitted frames and fragmented data streams. Future iterations of this framework will investigate advanced trajectory recovery mechanisms, such as integrating Kalman filtering with deep-learning-based re-identification (Re-ID) metrics or motion-guided spatial matching (e.g., historical velocity vector projection) and bio-inspired approaches [19]. By predicting missing positions and dynamically stitching fragmented fragments together, the system can preserve absolute identity continuity and further enhance the structural integrity of long-term biological swimming profiles.